\documentclass[10pt,twocolumn,letterpaper]{article}

\usepackage{cvpr}
\usepackage{times}
\usepackage{epsfig}
\usepackage{graphicx}
\usepackage{amsmath}
\usepackage{amssymb}
\usepackage{caption}
\usepackage{subcaption}
\usepackage{amsthm}
\usepackage[table,xcdraw]{xcolor}
\usepackage{multirow}
\usepackage{url}

\usepackage[pagebackref=true,breaklinks=true,letterpaper=true,colorlinks,bookmarks=false]{hyperref}

\cvprfinalcopy 

\def\cvprPaperID{1729} 

\ifcvprfinal\fi
\begin{document}

\title{Training Deeper Convolutional Networks with Deep Supervision}

\author{Liwei Wang\\
Computer Science Dept\\
UIUC\\
{\tt\small lwang97@illinois.edu}
\and
Chen-Yu Lee\\
ECE Dept\\
UCSD\\
{\tt\small chl260@ucsd.edu}
\and
Zhuowen Tu\\
CogSci Dept\\
UCSD\\
{\tt\small ztu0@ucsd.edu}
\and
Svetlana Lazebnik\\
Computer Science Dept\\
UIUC\\
{\tt\small slazebni@illinois.edu}
}

\maketitle

\begin{abstract}
One of the most promising ways of improving the performance of deep convolutional neural networks is by increasing the number of convolutional layers. However,  adding layers makes training more difficult and computationally expensive. In order to train deeper networks, we propose to add auxiliary supervision branches after certain intermediate layers during training. We formulate a simple rule of thumb to determine where these branches should be added. The resulting deeply supervised structure makes the training much easier and also produces better classification results on ImageNet and the recently released, larger MIT Places dataset.
\end{abstract}

\section{Introduction}




In the most recent ILSVRC competition~\cite{ILSVRCarxiv14}, it was demonstrated~\cite{simonyan2014very,szegedy2014going} that CNN accuracy can be improved even further by increasing the network size: both the depth (number of levels) and the width (number of units at each level). On the down side, bigger size means more and more parameters, which makes back-propagation slower to converge and prone to overfitting~\cite{simonyan2014very,szegedy2014going}. To overcome this problem, Simonyan and Zisserman~\cite{simonyan2014very} propose to initialize deeper networks with parameters of pre-trained shallower networks. However, this pre-training is costly and the parameters may be hard to tune. Szegedy et al.~\cite{szegedy2014going} propose to add auxiliary classifiers connected to intermediate layers. The intuitive idea behind these classifiers is to encourage the feature maps at lower layers to be directly predictive of the final labels, and to help propagate the gradients back through the deep network structure. However, Szegedy et al.~\cite{szegedy2014going} do not systematically address the questions of where and how to add these classifiers. Independently, Lee et al.~\cite{lee2014deeply} introduce a related idea of {\em deeply supervised networks} (DSN) where auxiliary classifiers are added at all intermediate layers and their ``companion losses'' are added to the loss of the final layer. They show that this deep supervision yields an improved convergence rate, but their experiments are limited with a not-so-deep network structure. 

In this work, to train deeper networks more efficiently, we also adopt the idea of adding auxiliary classifiers after some of the intermediate convolutional layers. We give a simple rule of thumb motivated by studying vanishing gradients in deep networks. We use our strategy to train models with 8 and 13 convolutional layers, which is deeper than the original AlexNet~\cite{krizhevsky2012imagenet} with 5 convolutional layers, though not as deep as the networks of~\cite{simonyan2014very,szegedy2014going}, which feature 16 and 21 convolutional layers, respectively. Our results on ImageNet~\cite{ILSVRCarxiv14} and the recently released larger MIT Places dataset~\cite{zhou2014places} confirm that deeper models are indeed more accurate than shallower ones, and convincingly demonstrate the promise of deep supervision as a training method. Compared to the very deep GoogleNet model trained on MIT Places dataset~\cite{googlePlaces}, an eight convolutional layer network trained with our method can give similar accuracy but with faster feature extraction. Our model on the Places Dataset is released in the Caffe Model Zoo~\cite{zoo}. 




\begin{figure*}
\centering \vspace{-0.2in}
 \includegraphics[width=1\textwidth]{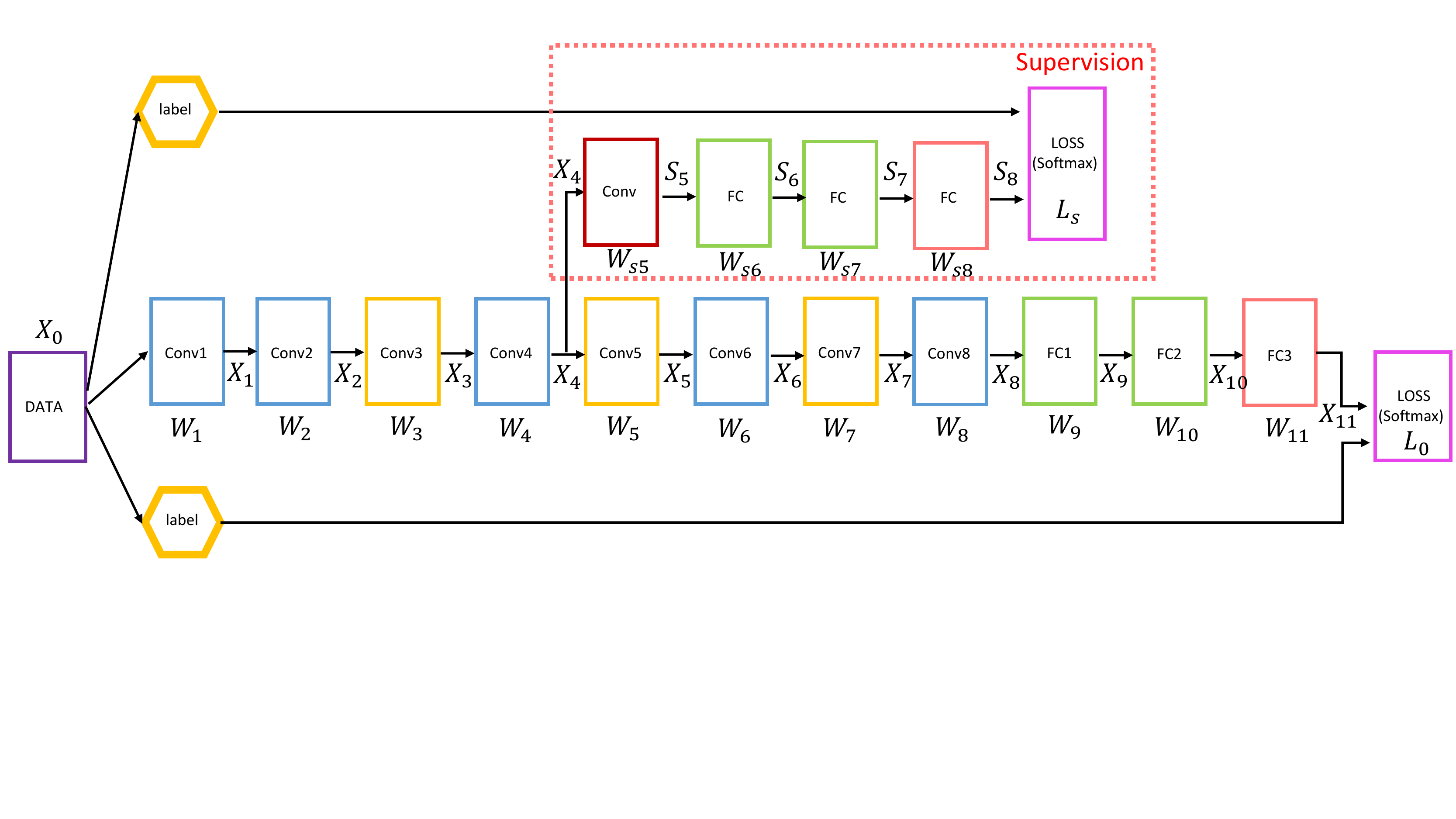} \vspace{-1.3in} \\
 \includegraphics[width=1\textwidth]{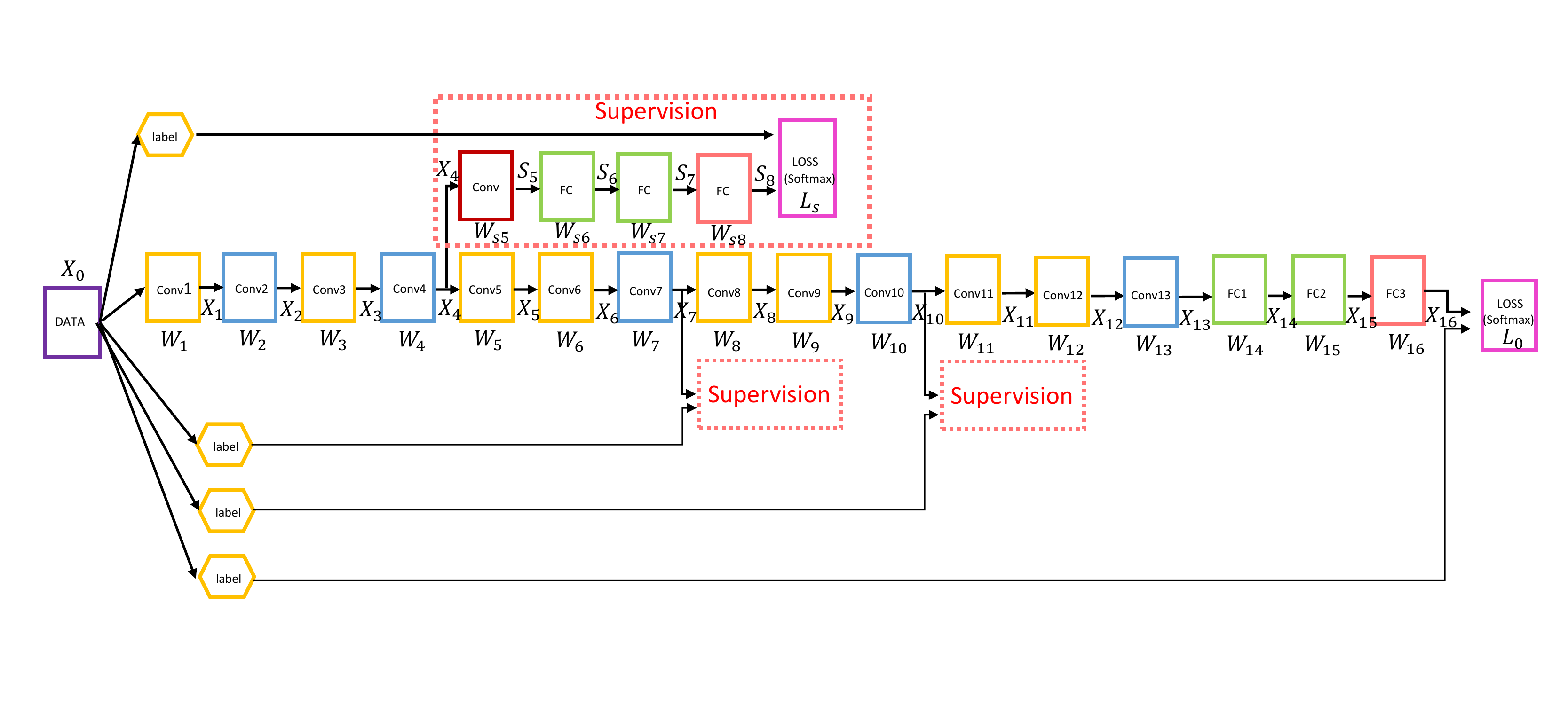} \vspace{-0.5in}  \\
\includegraphics[width=0.5\textwidth]{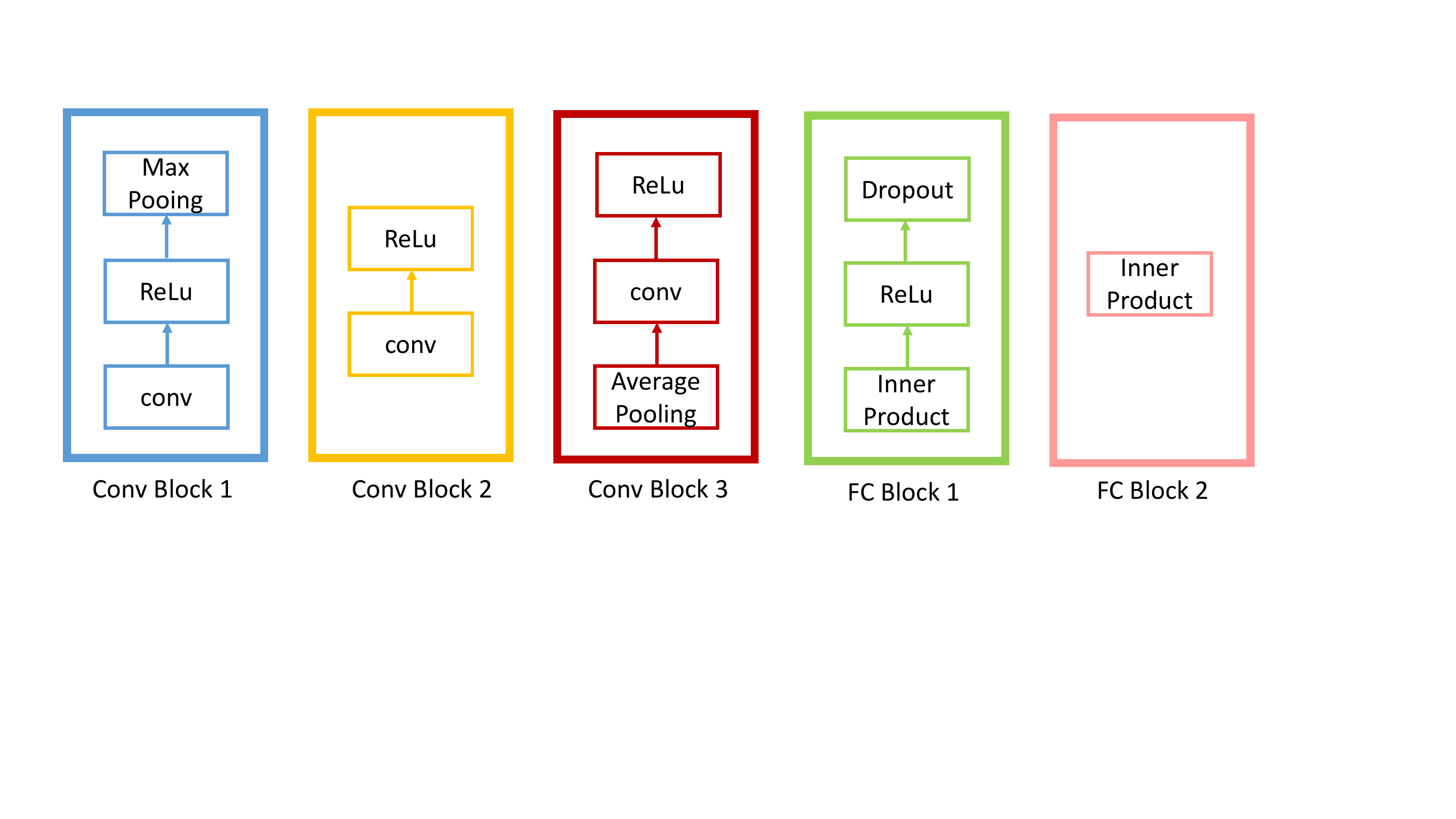}
  \vspace*{-0.5in}
  \caption{Illustration of our deep models with 8 and 13 convolutional layers. The additional supervision loss branches are indicated by dashed red boxes. $X_l$ denote the intermediate layer outputs and $W_l$ are the weight matrices for each computational block. Blocks of the same type are shown in the same color. A legend below the network diagrams shows the internal structure of the different block types.}\label{fig:BPplots}
\end{figure*}

\begin{figure*}
        \centering
        \begin{subfigure}[b]{0.40\textwidth}
                \includegraphics[width=\textwidth]{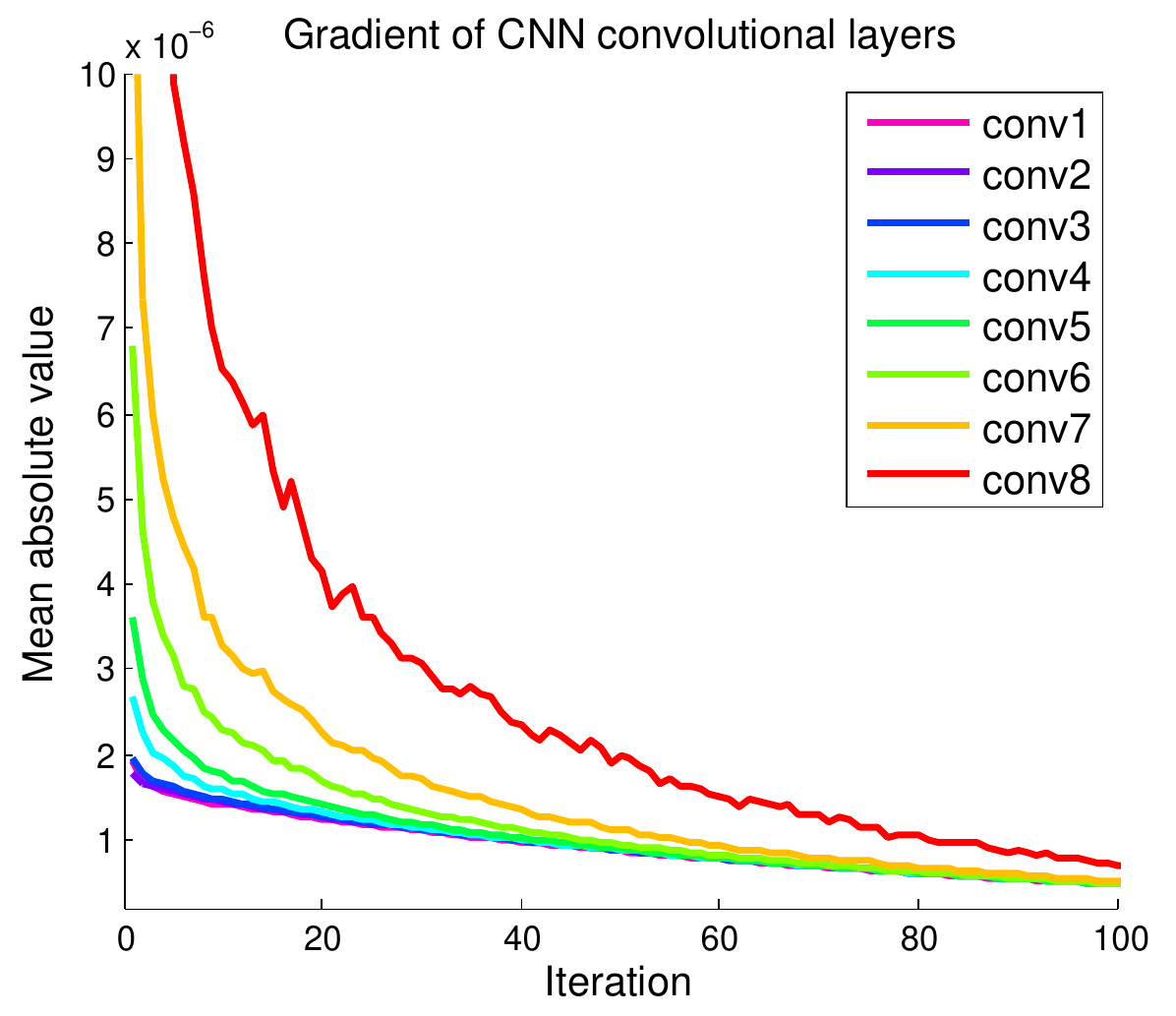}
                \caption{}
                \label{fig:gradient_CNN}
        \end{subfigure}%
        \begin{subfigure}[b]{0.45\textwidth}
                \includegraphics[width=\textwidth]{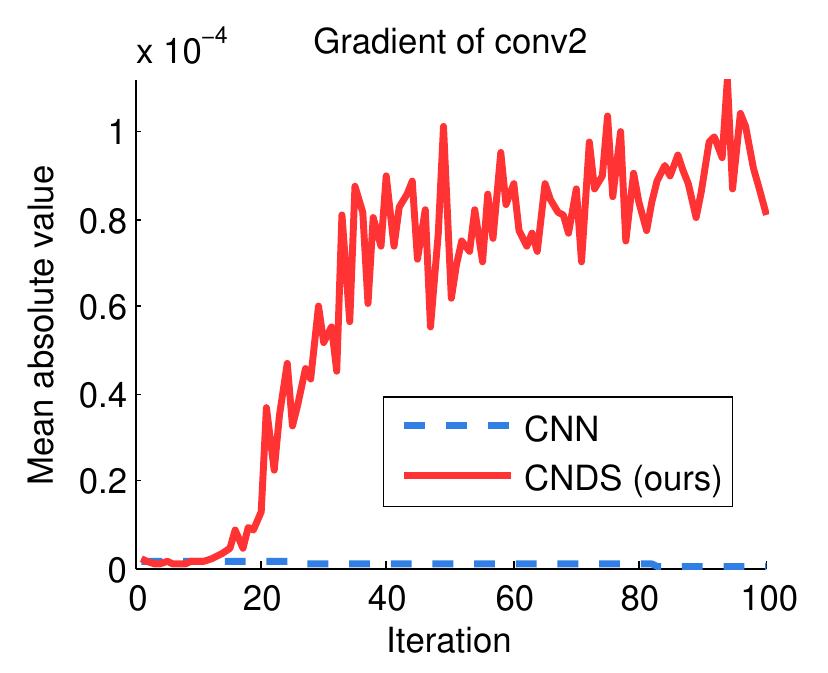}
                \caption{}
                \label{fig:conv3}
        \end{subfigure}
        \begin{subfigure}[b]{0.45\textwidth}
                \includegraphics[width=\textwidth]{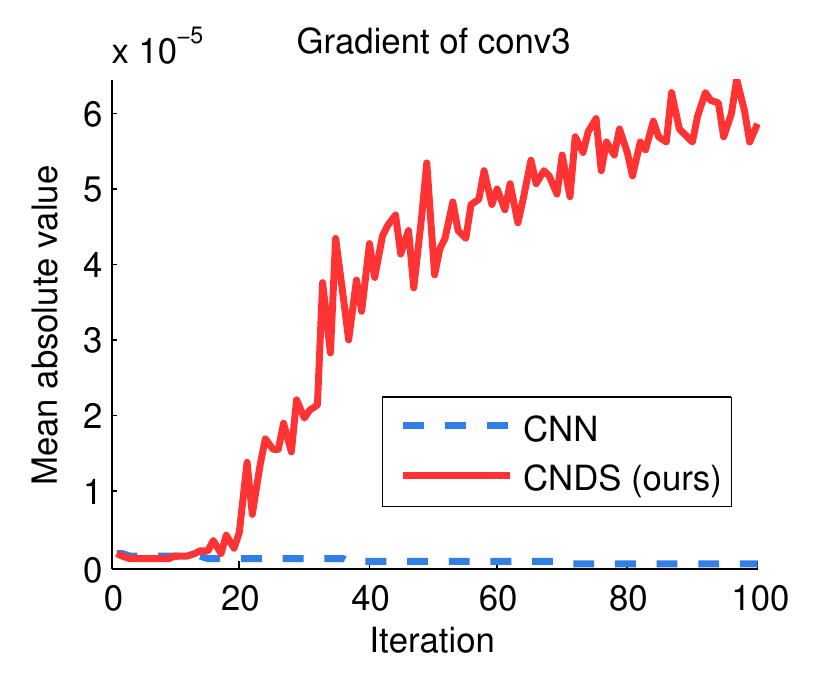}
                \caption{}
                \label{fig:conv4}
        \end{subfigure}
        \begin{subfigure}[b]{0.45\textwidth}
                \includegraphics[width=\textwidth]{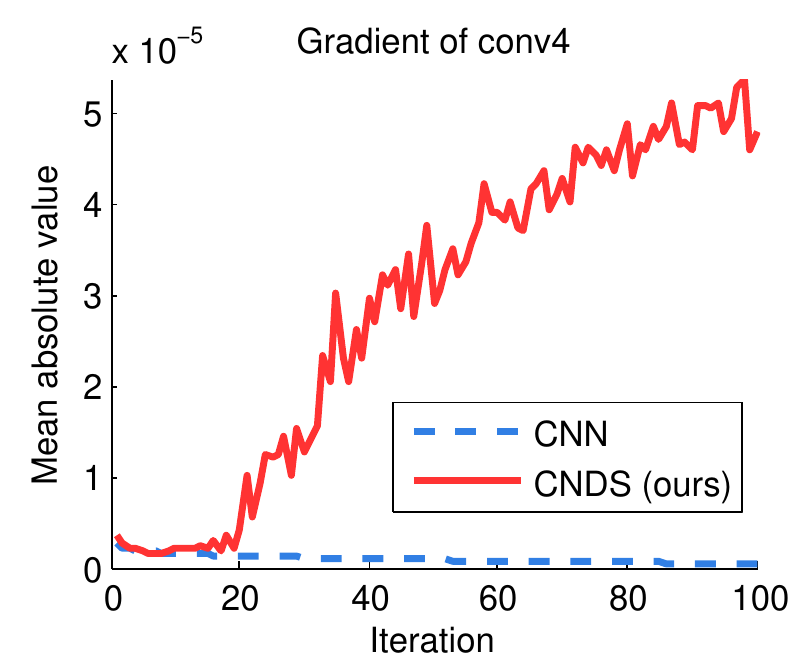}
                \caption{}
                \label{fig:conv4}
        \end{subfigure}
        \caption{(a) Mean gradient magnitude as a function of back-prop iterations for an 8-layer model trained on ImageNet with ``vanilla'' initialization (see text); (b-d)
        Comparison of gradient values before and after adding auxiliary supervision after the fourth convolutional layer. CNDS stands for our method: Convolutional Networks with Deep Supervision. For CNDS, after a few thousand iterations, the gradient magnitudes stop growing and stay steady for a long time.}\label{fig:gradient}
\end{figure*}

\section{Our Method} \label{sec:dsn}

Since very deep models have only made their d\'ebut in the 2014 ILSVRC contest, the problem of how to train them is just beginning to be addressed. Simonyan and Zisserman~\cite{simonyan2014very} initialize the lower convolutional layers of their deeper networks with parameters obtained by training shallower networks, and initialize the rest of the layers randomly. While they achieve very good results, their training procedure is slow and labor-intensive, since it relies on training models of progressively increasing depth, and may be very sensitive to implementation choices along the way. Szegedy et al.~\cite{szegedy2014going} connect auxiliary classifiers (smaller networks) to a few intermediate layers to provide additional supervision during training. However, their report does not give any general principles for deciding where these classifiers should be added and what their structure should be.

Lee et al.~\cite{lee2014deeply} give a more comprehensive treatment of the idea of adding supervision to intermediate network layers. Their {\em deeply-supervised nets} (DSN) put an SVM classifier on top of the outputs of each hidden layer; at training time, they optimize a loss function that is a sum of the overall (final-layer) loss and companion losses associated with all intermediate layers. 
Our work has a number of differences from~\cite{lee2014deeply}. First, they add supervision after each hidden layer, while we decide where to add supervision using a simple gradient-based heuristic described below. Second, the classifiers of~\cite{lee2014deeply} are SVMs with a squared hinge loss. By contrast, our supervision branch is more similar to that of~\cite{szegedy2014going} -- it is a small neural network composed of a convolutional layer, several fully connected layers, and a softmax classifier (see Fig.(\ref{fig:BPplots})). Since feature maps at the lower convolutional layers are very noisy, to achive good performance, we have found it important to put them through dimensionality reduction and discriminative non-linear mapping before feeding them into a classifier. 

To decide where to add the supervision branches, we follow an intuitive rule of thumb. First, we run a few (10-50) iterations of back-propagation for the deep model with supervision only at the final layer and plot the mean gradient values of intermediate layers (using the standard initialization for AlexNet~\cite{krizhevsky2012imagenet}: weights are sampled from a Gaussian with zero mean, std=0.01, and bias=0). Then we add supervision after the layer where the mean gradient value vanishes (in our implementation, becomes less than $10^{-7}$). As shown in Fig.(\ref{fig:gradient}), in our eight-layer model, the gradients in the fourth convolutional layer tend to vanish. Therefore, we add auxiliary supervision right after this layer.

The top of Fig.(\ref{fig:BPplots}) shows the resulting network structure, consisting of a main and an auxiliary supervision branch. In the main branch, weights $W_{1},...,W_{11}$ are associated with the eight convolutional and three fully connected layers. 
The auxiliary branch comes with weights $W_{s5},W_{s6},W_{s7},W_{s8}$.
Let $\mathcal{W}$ and $\mathcal{W}_s$ denote the concatenations of the two respective sets of parameters:
\begin{eqnarray} \label{eqn:loss}
\mathcal{W}&=& (W_{1}, \ldots, W_{11})\,, \nonumber \\
\mathcal{W}_{s} &=& (W_{s5}, \ldots, W_{s8}) \,. \nonumber
\end{eqnarray}

Given a training example that produces feature map $X_{11}$ at the softmax layer, for each possible label
$k = 1, \ldots, K$ this layer computes the response
\[ p_{k} =  \frac{\exp(X_{11(k)})}{\sum_{k} \exp(X_{11(k)})} \,, \]
where $X_{11(k)}$ is the $k$ th element of response $X_{11}$.
The associated loss for the entire network is
\[ \mathcal{L}_{0}(\mathcal{W}) = - \sum_{k=1}^K y_{k} \ln p_{k}\, \]
where $y_k = 1$ if the example has label $k$ and 0 otherwise.

Analogously, given feature map $S_8$ before the softmax layer of the auxiliary supervision branch, we have the output
\[ p_{sk} = \frac{\exp(S_{8(k)})}{\sum_{k} \exp(S_{8(k)})} \]
where $S_{8(k)}$ is the $k$ th element of response $S_{8}$
and associated auxiliary loss
\[ \mathcal{L}_{s}(\mathcal{W},\mathcal{W}_s) = - \sum_{k=1}^K y_{k} \ln p_{sk} \, .\]
Note that this loss depends on $\mathcal{W}$, not just $\mathcal{W}_s$, because the computation of the feature map $S_8$
involves the weights of the early convolutional layers $W_1, \ldots W_4$.

The combined loss function for the whole network is given by a weighted sum of the main loss $\mathcal{L}_{0}(\mathcal{W})$ and
the auxiliary supervision loss $\mathcal{L}_{s}(\mathcal{W}_{s})$:
\begin{eqnarray} \label{eqn:loss}
\mathcal{L}(\mathcal{W},\mathcal{W}_{s}) = \mathcal{L}_{0}(\mathcal{W}) + \alpha_{t} \mathcal{L}_{s}(\mathcal{W},\mathcal{W}_{s}) \,,
\end{eqnarray}
where $\alpha_{t}$ controls the trade-off between the two terms. In the course of training, in order to use the second term mainly as regularization, we adopt the same strategy as in~\cite{lee2014deeply}, where $\alpha$ decays as a function of epoch $t$ (with $N$ being the total number of epochs):
\begin{equation} \label{eq:alpha}
\alpha_{t} \leftarrow \alpha_{t}*(1-t/N)\,.
\end{equation}



We train our deeply supervised model using stochastic gradient descent. When doing back-propagation, $W_{5},\ldots,W_{11}$, are only affected by the main loss $\mathcal{L}_0$. Similarly, $W_{s5},\ldots,W_{s8}$ are only affected by the auxiliary loss $\mathcal{L}_{s}$. However, starting from $W_{4}$, where the gradients tend to vanish, the $\mathcal{L}_{s}$ term successfully magnifies the gradients, as can be seen from the before-and-after comparisons in Fig.(\ref{fig:gradient})(b-d).

In addition to our 8-layer model, we have also experimented with a 13-layer model (Fig.(\ref{fig:BPplots}), middle). For this model, gradients tend to decay every three to four layers, and we get good results by putting the supervision branches after the 10th, 7th, and 4th layers. All the auxiliary losses have the same weights $\alpha_t$ (starting with 0.3 and then decaying according to eq. (\ref{eq:alpha})). We do not give the resulting loss functions here, but their derivation is straightforward.

In the following, we will refer to our training method as CNDS (Convolutional Networks with Deep Supervision).

\section{Experiments}

Sections \ref{sec:imagenet} and \ref{sec:places} will present an evaluation of our models trained on the two largest datasets currently available: ImageNet (ILSVRC)~\cite{ILSVRCarxiv14} and MIT Places~\cite{zhou2014places}.


\subsection{ImageNet Results} \label{sec:imagenet}

We report results on the ILSVRC subset of ImageNet~\cite{imagenet}, which includes 1000 categories and is split into 1.2M training, 50K validation, and 100K testing images (the latter have held-out class labels). The classification performance is evaluated using top-1 and top-5 classification error.
Top-5 error is the proportion of images such that the ground-truth class is not within the top five predicted categories.

All our ImageNet models are trained with Cuda-convnet2 (https://code.google.com/p/cuda-convnet2/) using epoch unit. Because Cuda-convnet2 supports multi-GPU training, we can train deeper networks in a reasonable time. We use Caffe~\cite{jia2014caffe} default setting for training ImageNet, that is, we crop one center and four corner patches of size $227\times227$ pixels (out of $256\times256$) and do horizontal flipping. We {\em do not} use model averaging or multi-scale training/testing. Please see the supplementary material for details of all the implementation settings for our models.

First, we survey the top systems in the ILSVRC competitions from 2012 to 2014. For models with five convolutional layers, in the 2012 version of the contest, the highest results were achieved by Krizhevsky et al.~\cite{krizhevsky2012imagenet}, who have reported 40.7\% top-1 and 18.2\% top-5 error rate using a single net. Subsequently, Zeiler and Fergus~\cite{zeiler2014visualizing} have obtained 36.0\% top-1 and 16.7\% top-5 error rates by refining Krizhevsky's filters and combining six nets. In the 2013 competition, Sermanet et al.'s OverFeat~\cite{sermanet2013overfeat} obtained 34.5\% top-1 and 13.2\% top-5 error rate by combining seven nets. In the 2014 competition, big progress was made by deeper models: the VGG group~\cite{simonyan2014very} has trained a series of deep models to get 25.5\% top-1 and 8.0\% top-5 error. And a 22-layer GoogLeNet~\cite{szegedy2014going} has achieved a top-5 error rate of 6.7\%. For a fair comparison, Table \ref{table:ImageNet}(a) lists single-model results from each of these systems.

In this work, we train networks with 8 and 13 convolutional layers using deep supervision. First, as a baseline, we train an ImageNet-CNN-8 model, which contains 8 convolutional layers and 3 fully connected layers, using the strategy in~\cite{simonyan2014very}: we first train a network with five convolutional layers, and then we initialize the first five convolutional layers and the last three fully connected layers of the deeper network with the layers from the shallower network. The other intermediate layers are initialized randomly. Including the time for training the shallower network, ImageNet-CNN-8 takes around 6 days with 80 epochs on two NVIDIA Tesla K40 GPUs with batch size 128.

Next, we train an ImageNet-CNDS-8 model using our deep supervision method. This model is trained with auxiliary supervision added after the fourth convolutional layer as shown in Fig.(\ref{fig:BPplots}). This model takes around 5 days to train with 65 epochs on two K40 GPUs with batch size 128. The weight $\alpha_{t}$ starts with 0.3 in all our CNDS training and decays according to eq. (\ref{eq:alpha}). Apart from the initialization and training procedure, we use the same network and parameter settings for both ImageNet-CNDS-8 and ImageNet-CNN-8.
It should be noted that in the testing phase, the auxiliary supervision branch of the CNDS model is cut off so it has the same feedforward path as the CNN model.

In order to go deeper, we also train an ImageNet-CNDS-13 model with structure shown in Fig. (\ref{fig:BPplots}). ImageNet-CNDS-13 takes around 5 days on {\em four} GPUs using 67 epochs with batch size 128. Since initializing weights for such a deep structure is in itself an open problem, we only train it with our CNDS method.

Table \ref{table:ImageNet}(b) shows the top-1 and top-5 accuracies of our models on the validation set of ILSVRC. First, both our 8-layer models outperform state-of-the-art 5-layer models from the literature, and the 13-layer model outperforms the 8-layer models. Therefore, ``going deeper'' really is an effective way to improve classification accuracy. Second, ImageNet-CNDS-8 is 1\% more accurate than ImageNet-CNN-8, while taking less time to train. It is important to emphasize that both models have the exact same structure at test time. Therefore, we can think of deep supervision as a form of regularization that gives better local minima for the classification task (since $\alpha_t$ eventually decays to zero, at the end, the loss we are optimizing is the original loss $\mathcal{L}_0$).

In absolute terms, our models are still less accurate than the VGG models of the same depth. However, we believe that this difference mainly comes from the network settings. In particular, we use a stride of 2 and filter size of 7 at the first layer, while they use a stride of 1 and filter size of 3, which gives them a finer-grained representation; we use single-scale training, while they use multi-scale. However, all other factors being equal, deep supervision may be a more promising strategy for training very deep networks than the iterative deepening scheme of~\cite{simonyan2014very}, since it is less complex and takes less time to train.

\subsection{Places Results} \label{sec:places}

ImageNet images mainly have center around objects, while the recently released MIT Places dataset~\cite{zhou2014places} is scene-centric. For training of deep networks, a subset of Places, called Places-205, has been created, which contains 2.4M training images from 205 categories, with 5000-15000 images per category. The validation set contains 100 images per category and the test set contains 200 images per category (with held-out class labels). The training set of Places is almost two times larger than the ILSVRC training set and 60 times larger than the SUN dataset~\cite{xiao2010sun}.

As a baseline, we use the five-layer net that was released along with Places. It was trained using the Caffe package on a GPU NVIDIA Tesla K40. As reported in~\cite{zhou2014places}, the process took 6 days and 300K iterations. We train two models for comparison: Places-CNN-8 and Places-CNDS-8, whose structure and parameters are the same as those of ImageNet-CNN-8 and ImageNet-CNDS-8. The only difference is that we train these models using Caffe instead of Cuda-convnet2, to stay consistent with the pre-trained Places model. Places-CNN-8 is trained the same way as ImageNet-CNN-8, using a pre-trained five-layer network as initialization. Including the pre-training time, Places-CNN-8 takes around 8 days with 240,000 iterations on a single K40 GPU with batch size 256 (Caffe only allows single-GPU training). Places-CNDS-8 takes around 6 days with 190,000 iterations. Same as for ImageNet, the weight $\alpha_{t}$ starts with 0.3 and decays according to eq. (\ref{eq:alpha}).

Following the evaluations in \cite{zhou2014places}, we give the top-1 and top-5 {\em accuracy} on both validation and test set. Note that for the test set, the ground-truth labels are not released. Instead, we sent our the prediction of both top 1 and top 5 labels to the MIT testing website (http://places.csail.mit.edu/submit.html) and got the results automatically. 

From Table (\ref{top5_places}), we can see that Places-CNN-8 and Places-CNDS-8 both outperform the original Places-CNN-5. Consistent with the results on ImageNet, our CNDS model surpasses Places-CNN-8 by about 1\%. Overall, with a combination of a deeper model and deep supervision, we achieve about 5\% higher accuracy than the baseline number reported in~\cite{zhou2014places}. Our network compares favorably with the one trained with the GoogleNet structure, and has faster feature extraction speed since it is not as deep.

%

\begin{table}
\begin{center}
\begin{tabular}{llccc}
\hline
& Method & layers  & top-1 & top-5 \\
\hline
(a) & Krizhevsky et al.~\cite{krizhevsky2012imagenet} & 5 & 40.7  & 18.2  \\
& OverFeat~\cite{sermanet2013overfeat} & 5   & 39.0  & 16.9 \\
& Zeiler and Fergus~\cite{zeiler2014visualizing} & 5  & 37.5   & 16.0  \\
& VGG~\cite{simonyan2014very} & 8  & 29.6   & 10.4 \\
& VGG~\cite{simonyan2014very} & 13  & 27.0  & 8.8 \\
& GoogLeNet~\cite{szegedy2014going} & 21  & - & 10.07 \\
\hline
(b) & ImageNet-CNN-8 & 8 & 34.7  & 14.0 \\
& ImageNet-CNDS-8 & 8 & 33.8  & 13.2 \\
& ImageNet-CNDS-13 & 13 & 31.8  & 11.8 \\
\end{tabular}
\end{center}
\caption{ILSVRC 2012-2014 results: top-1 and top-5 {\em error} on validation set.
(a) Top single-model results from the literature (see text for additional results and discussion).
(b) Results for our models.}
\label{table:ImageNet}
\end{table}

\begin{table}
\begin{center} 
\begin{tabular}{l*{6}{c}r}
\hline
Methods     & top-1 val/test & top-5 val/test \\
\hline
Places-CNN-5 \cite{zhou2014places} & 50.4 / 50.0   &  80.9 / 81.1 \\
Places-GoogleNet \cite{googlePlaces} & -/ 56.3    & -/ 86.0 \\
Places-CNN-8 (ours) & 54.0 / 53.8 & 83.7 / 83.6 \\
Places-CNDS-8 (ours) & \textbf{54.7} / \textbf{55.7}  & \textbf{84.1} / \textbf{85.8} \\
\end{tabular}   
\end{center}
\caption{Top-1 and top-5 {\em accuracies} on Places validation and test sets for the 5-layer model from~\cite{zhou2014places} and our 8-conv layer models.}
\label{top5_places}
\end{table}

\section{Discussion}

This work has focused on the idea of training very deep networks with auxiliary supervision inserted at intermediate layers. We have attempted to formulate sound design principles of where and how deep supervision should be added. Our experiments have also shown the advantage of this technique over alternative methods that require pre-training of shallower networks~\cite{simonyan2014very}. Along the way, we have reported new state-of-the-art results on the recently released very large Places dataset~\cite{zhou2014places}.

{\small
\bibliographystyle{ieee}
\bibliography{egbib}
}

\end{document}